# Robust imaging of hippocampal inner structure at 7T: in vivo acquisition protocol and methodological choices

Linda Marrakchi-Kacem[1,2,3,4,5,6*], Alexandre Vignaud[7], Julien Sein[8], Johanne Germain[1,2,3,4,5], Thomas R. Henry[9], Cyril Poupon[7], Lucie Hertz-Pannier[10,11,12], Stéphane Lehéricy[1,2,3,4,13], Olivier Colliot[1,2,3,4,5], Pierre-François Van de Moortele[8], Marie Chupin[1,2,3,4,5]

1 Sorbonne Universités, UPMC Univ Paris 06, UMR S 1127, ICM, F-75013, Paris, France
2 Inserm, U1127, F-75013, Paris, France
3 CNRS, UMR 7225, F-75013, Paris, France
4 ICM, 75013, Paris, France
5 Inria Paris-Rocquencourt, 75013, Paris, France
6 Université de la Manouba, Institut Supérieur de Biotechnologies de Sidi Thabet, Tunis, Tunisie
7 UNIRS , NeuroSpin, I2BM, DSV, CEA-Saclay, France
8 Center for Magnetic Resonance Research (CMRR), University of Minnesota, Minneapolis, MN, USA
9 Department of Neurology, University of Minnesota, Minneapolis, MN, USA
10 UNIACT , NeuroSpin, I2BM, DSV, CEA-Saclay, France
11 INSERM U1129, Paris, France;
12 Paris-Descartes university; CEA, Gif sur Yvette, France
13 Centre de Neuro-Imagerie de Recherche CENIR, AP-HP, Hopital de la Pitie Salpetriere, Paris, France

**\*corresponding author**
tel: +21696313381
Email: linda.marrakchi@gmail.com

Abstract word count: 193

Text word count: 7840

Figures number: 12

Table number: 4

References number: 34

**Abstract**

*Objective:* Motion-robust multi-slab imaging of hippocampal inner structure in-vivo at 7T.

*Material and methods:* Motion is a crucial issue for ultra-high resolution imaging, such as can be achieved with 7T MRI. An acquisition protocol was designed for imaging hippocampal inner structure at 7T. It relies on a compromise between anatomical details visibility and robustness to motion. In order to reduce acquisition time and motion artefacts, the full slab covering the hippocampus was split into separate slabs with lower acquisition time. A robust registration approach was implemented to combine the acquired slabs within a final 3D-consistent high-resolution slab covering the whole hippocampus. Evaluation was performed on 50 subjects overall, made of three groups of subjects acquired using three acquisition settings; it focused on three issues: visibility of hippocampal inner structure, robustness to motion artefacts and registration procedure performance.

*Results:* Overall, T2-weighted acquisitions with interleaved slabs proved robust. Multi-slab registration yielded high quality datasets in 96% of the subjects, thus compatible with further analyses of hippocampal inner structure.

*Conclusion:* Multi-slab acquisition and registration setting is efficient for reducing acquisition time and consequently motion artifacts for ultra-high resolution imaging of the inner structure of the hippocampus.

1. Introduction

The hippocampus is a complex brain structure located in the medial part of the temporal lobe. It plays a crucial role in episodic and working memory, emotional processes and memory consolidation [1–3]. Its involvement has been extensively studied in several neuropsychiatric disorders, such as Alzheimer disease[4–7] and temporal lobe epilepsy [8]. Hippocampal subparts have been proven diversely affected by pathological processes, and in-vivo imaging of these subparts is thus of high interest to study these disorders [9]. High field MR imaging made it possible to visualize hippocampal inner structure in-vivo, with 3T [10, 11], 4.7T [12–17] and 7T [18–21] MR systems. Ultra-high field imaging (7T and above) provided both better contrast to noise ratio and higher spatial resolution and made it possible to distinguish between the layers with lower and higher density of neuronal bodies [18, 19, 21–26].

Previous work with in-vivo 7T imaging of the hippocampus has mostly focused either on assessing hippocampal inner structure visibility in healthy subjects [18, 19, 22] or on designing manual segmentation protocols for hippocampal inner structure [20, 21]. Few studies have investigated alterations of hippocampal inner structure in patients [23–27]. Acquisition times mostly ranged from 10 [20, 24] to 15 minutes [19] with various resolutions, contrasts and fields of view, for either 2D or 3D acquisitions. While such acquisition times could be feasible for healthy subjects, they are likely to increase the frequency of motion during acquisitions for patients. Motion during acquisitions will typically yield blurring or ghosting artefacts in the hippocampal area, thus either reducing hippocampal inner structure visibility or creating spurious edges mimicking hippocampal inner structure. In order to reduce the duration of each individual scan, and thus reduce motion likelihood, we previously proposed to split the ten minutes long scan in two separate five minutes long scans [23]. 2D acquisitions can easily be split by acquiring two slabs (stacks of slices), each one embedding half the final number of slices; these slabs can be spatially distributed either in two adjacent slabs with contiguous slices or in two interleaved slabs with a gap between slices equal to the slice thickness. With this approach, however, subjects' movement between acquisitions is likely to yield between-slab shift. This kind of approach thus requires a specific slab acquisition and registration procedure in order to build a 3D-consistent stack of slices.

In this work, we propose a complete acquisition procedure that relies on multi-slab acquisitions together with a dedicated registration framework to obtain 3D-consistent ultra-high resolution images of the whole hippocampus at 7T. Before describing the multi-slab acquisition procedure and registration method, details are given below regarding the characteristics of the inner structure of the hippocampus. In fact, hippocampal inner structure is anatomically characterized by an intricate set of very small subparts. In order to make it possible to analyze this inner structure, two criterions are to be taken into account: the size of the structure that will determine spatial resolution and the geometry that will allow to derive simplifying procedures.

1.1 Anatomy

The hippocampus can be divided into three main parts: the head (anterior part), the body (middle part) and the tail (posterior part) (Figure1.a). In healthy adults, its total length ranges from 40 to 45mm and its height is about 7mm. Its width ranges from about 10mm in the body and 15 to 20mm in the head [28].

The hippocampus is histologically defined as a bilaminar structure: two thin substructures, the cornu Ammonis (CA) and the gyrus dentatus (GD), are wrapped around one another (Figure1.b). The thickness of CA ranges approximately from 1 to 1.5mm. It can be subdivided into six layers characterized by different cellular content (figure1.c): alveus, stratum oriens, stratum pyramidale, stratum radiatum, stratum lacunosum and stratum moleculare [28]. Neuronal bodies are closely packed in the stratum pyramidale (SP) and sparse in the stratum radiatum (SR), stratum lacunosum (SL), and stratum moleculare (SM). The alveus is made of axons whereas the stratum oriens contains both neuronal bodies and axons. These differences are assumed to be a main underlying source of MR contrast, allowing for the distinction between three groups of layers in high resolution MR images of the hippocampus: SRLM (SL, SR, SM), SP and alveus (alveus and stratum oriens).

1.2 MR contrast and hippocampal inner structure visualization

The visualization of hippocampal inner structure at 7T has mainly been investigated with three MR contrasts: T1, T2 and T2* [18–24]. Few studies considered T1-weighting for visualizing hippocampal inner structure, either at 7T [18, 22] or at lower field strengths [11, 29]. T2-weighting has been the most widely used MR contrast for hippocampal inner structure analyses [20, 22–26], as it makes it possible to clearly distinguish between SRLM, SP and alveus groups; the lower signal in the SRLM and the alveus yields a big contrast with the higher signal in both the SP and the GD. T2*-weighting was also used at 7T to visualize the hippocampal inner structure [18, 19, 21, 22, 30], with sometimes even higher contrast between the SRLM and the SP than with T2-weighting. However, T2*-weighted acquisitions are very sensitive to susceptibility-induced artifacts, in particular in the temporal lobe region due to the proximity of the ear canals. Both T2- and T2*-weightings were evaluated in this study.

1.3 Geometry: spatial resolution and acquisition plane

The hippocampus is also characterized by its antero-posterior span; in fact its main axis is close to sagittal (Figure1.a) and the geometry of its inner structure appears more stable and better defined perpendicular to it, especially in the body where CA and GD wrap around one another along it (Figure1.b). The first protocols proposed for analyzing the hippocampus were thus designed for coronal slices acquired perpendicular to its main axis, through a tilt of either the subject's head in the MR system or the acquisition slab [31]. However, imaging hippocampal inner structure remains highly challenging because of the size of its subparts as distinguished with MRI: the thickness of the two major ribbon-shaped groups of layers, SP and SRLM, ranges from 0.5 (or even less) to 1mm in healthy adults, with large variations along CA. In order to visualize and precisely measure SP and SRLM in both healthy subjects and patients, resolution of about 200 μm to 400 μm is necessary.

Unfortunately, acquiring sufficient signal with such an isotropic high resolution is not currently feasible with standard sequences in realistic acquisition times. This would suggest using 3D sequences; however,

3D T2-weighted sequences, with standard repetition time, would be far too long to be feasible in patients. Specific 3D T2-weighted sequences have been developed, e.g. sequences exploiting stimulated echoes[32], which are much faster to acquire but require complex modulations of variable flip angle and generally yield a mix of T1, proton density and T2 weighting that are sensitive to transmit B1 inhomogeneities observed at 7T [33, 34].

Taking advantage of the anisotropic organization of hippocampal inner structure, an efficient compromise between resolution and feasibility can be obtained with standard 2D coronal oblique sequences, acquired perpendicular to the main axis of the hippocampus, with in-plane resolution of about 250 - 300 µm and slice thickness of about 1 – 1.2 mm. Previous publications have shown that such acquisitions can provide sufficient visibility of hippocampal inner structure [23] with suitable acquisition times for clinical setting. However, imaging the full hippocampus with such geometrical characteristics still requires about ten minutes, making motion more likely to occur during acquisitions. In order to maintain high in-plane resolution while decreasing the probability of motion artifacts, the slab required to cover the full extent of hippocampus can be acquired in multiple slabs.

2. **Material and methods**

Between-slab shift is a crucial issue for multi-slab acquisitions. However, thanks to the close fitting geometry of the coil, head movements are very restricted and some shifts are not even possible. Besides the hippocampal area will not be affected in a similar way by all the remaining possible movements. A specific registration-based correction for between-slab shifts is necessary in order to build a 3D-consistent stack of slices.

2.1 Between-slab shift configurations

Multi-slab acquisition consists in acquiring n complementary slabs and can be carried out in two ways: either with contiguous (or slightly overlapping) slabs (Figure 2.a) or with interleaved slabs (Figure 2.b) sequentially acquired to cover the full extent of the hippocampus.

The consequences of possible head movements between complementary slab acquisitions was considered (Figure 3) with a special focus on possible partial information loss. In fact, some motion configurations could result in a local between-slab translation along the main axis of the hippocampus: part of the signal from the first slab would also be acquired in the complementary slab while part of the signal from that complementary slab would be missing. While such a pure translation along the main axis of the hippocampus is very unlikely (Figure 3.e), a rotation around the left-right axis (Figure 3.c) or the superior-inferior axis would yield local movement along the main axis of the hippocampus. In case of contiguous complementary slabs, this issue would involve few voxels only at the interface between these complementary slabs; it could easily be addressed by adding some overlapping slices at this interface. On the contrary, in case of interleaved complementary slabs, information loss would involve voxels throughout the whole slab. Fortunately, the hippocampus being located close to the center of the head, it is likely to be in an area with limited consequences of between-slab movement along the main axis of the hippocampus, and it should not be altered by most realistic rotations.

2.2 3D-consistent slab reconstruction

The aim of the procedure was to combine the acquired slabs into a single 3D-consistent high resolution volume. For both contiguous and interleaved slabs, the multi-slab registration procedure was made more robust by the acquisition of an additional lower resolution (LR) continuous slab that covered the whole length of the hippocampus and that was used as a target for registration. Consistency for final intensity values was ensured through the use of binary masks mimicking the signal content of the acquired slabs.

The high resolution (HR) slabs were acquired with an in-plane resolution of $r_x*r_z$ and a slice thickness of $r_y$. Here, two slabs made of N slices each were considered. Two configurations were considered:

A) Contiguous slabs: the two slabs cover the anterior and posterior parts of the hippocampus respectively, with one overlapping slice;
B) Interleaved slabs: each slab covers the whole hippocampus and contains N slices with 100% gap.

In order to register each HR slab to the LR volume, three preliminary steps were performed:

- The LR volume was resampled to $r_x*r_z$ in-plane resolution;
- Each HR slab was completed with slices with no signal ("null slices") to reach the dimension of the final volume:
    A) Contiguous slabs: N null slices were added to each HR acquired slab, corresponding to the complementary slab (e.g. the null slices were added posterior to the anterior acquired slab);
    B) Interleaved slabs: N null slices (corresponding to the gaps between slices) were inserted between the slices with acquired signal.
- For each HR slab, a mask was created with $r_x*r_z$ in-plane resolution and a slice thickness of $r_y$ with the same matrix size as the HR slab: voxels were set to 1.0 for slices with acquired signal and to 0.0 for the null slices.

The procedure was implemented with the pyaims library (http://brainvisa.info/index_f.html). Each HR volume was registered to the LR volume using the SPM8 (http://www.fil.ion.ucl.ac.uk/spm/) coregistration module which performs a rigid transform based on normalized mutual information criterion and with B-spline interpolation method. The associated masks were also registered to the LR volume with the same transformation matrix as their corresponding HR volumes. The final full HR volume was obtained by dividing the sum of the registered HR volumes by the sum of the registered masks; this step ensured consistent intensities in the final slab even for partly overlapping voxels. The registration pack is available for download from the following website: http://www.aramislab.fr/sevenhipp.

2.3     MR acquisition

MR acquisitions were performed in two research centers (NeuroSpin, CEA, France and Center for Magnetic Resonance Research (CMRR), University of Minnesota, MN, USA) on three 7T acquisition settings as described in Table 1. The respective regional ethics committees approved the studies and written informed consent was obtained from all participants. The characteristics of the sequences are described below and the acquisition parameters are summarized in Table 2. Some sequences were acquired in all subjects while others were acquired only for a subset of subjects, as they were discarded as not useful after a first series of analysis described in the result section. Some sequences were acquired twice or more to evaluate motion evolution during the scanning session and/or the advantages of averaging for better visibility. The standard protocol is described in Table 3. Manual shim procedures were applied together with the use of the Siemens correction of the distortions due to gradient non-linearity. Coronal oblique slabs were positioned perpendicular to the main axis of the hippocampus; in NeuroSpin, a specific localizer built with two sagittal slabs centered on temporal lobes was acquired to facilitate the prescription of the coronal oblique views. In CMRR, The coronal oblique orientation was prescribed from the 3D T1w MPRAGE images.

**T2-weighted acquisitions:** Slightly different 2D Turbo Spin Echo (TSE) sequences were acquired on NS_7T_32CH, CMRR_7T_16CH and CMRR_7T_32CH with 0.25 to 0.3mm in-plane resolution and 1.2mm slice thickness (Table 2). Both contiguous and interleaved acquisitions were investigated, with an acquisition time of about five minutes per slab. A LR slab was also acquired with a voxel size twice larger than for the HR acquisition in the phase encoding direction, thus making it possible to acquire the full extent of the hippocampus in about five minutes; voxels were interpolated to in-plane isotropic resolution during reconstruction. Note that, due to SAR issues, the hippocampus was acquired with four slabs for the CMRR_7T_32CH datasets: two interleaved slabs covered the anterior part of the hippocampus, and two interleaved slabs covered the posterior part of the hippocampus, with an overlapping slice between anterior and posterior slabs.

**T2\*-weighted acquisitions**: A 2D Gradient Echo (GRE) sequence was acquired with three interleaved slabs, 0.3*0.3mm in plane resolution and 1.2mm slice thickness, either as three separate acquisitions of about four minutes each or in a single acquisition of about 12 minutes.

2.4 Evaluation procedure

Visual assessment was performed by a rater trained in the anatomy of hippocampal inner structure. The aim was to evaluate hippocampal inner structure visibility, prevalence of motion artifact and its evolution during the acquisition session, performance of the between-slab registration procedure and prevalence of between-slab information loss.

2.4.1 Visibility of hippocampal inner structure

The visibility of hippocampal inner structure was first evaluated overall for T2-weighted TSE and T2*-weighted GRE acquisitions performed with the three acquisition settings in order to assess: 1. how T2* weighting compared with the more standard T2 weighting for analyzing hippocampal inner structure; 2. whether the T2-weighted images acquired with the three acquisition settings were suitable for such analysis (with and without averaging). Visibility was defined in the acquisition plane as the possibility to differentiate between the SP, the SRLM and the alveus in the head, the body and the anterior part of the tail. The most posterior slices were not taken into account because of a sharp tilt of the main axis of the hippocampus in these slices; in fact, the change of main orientation made the discrimination more difficult in oblique coronal slices perpendicular to the main axis of the hippocampus (note that this part was not studied in detail in previous segmentation protocols [12, 17, 20, 29]). Some ambiguous slices may remain but they should be few enough to be understood by taking into account the other slices. Subjects with large motion artifacts were not taken into account for this evaluation.

A systematic comparison was performed by two raters in order to visually compare T2-weighted TSE contiguous and interleaved HR slabs. For each subject, each rater rated interleaved and contiguous acquisitions according to its visibility of the inner structure of the hippocampus with three level scale (0 if worst visibility, 0.5 if equal visibility, 1 if best visibility). Images were first checked for motion artifacts, and only subjects for whom both acquisitions were without motion artifacts were kept, in order to study acquisition quality in itself without being biased by sensitivity to motion.

Relative contrast (RC) and signal to noise ratio (SNR) were also computed as follows for a representative subject. Ellipsoid were manually placed following a systematic procedure in the same anatomical regions for all images: one for white matter (WM) in an homogeneous subcortical area in the right hemisphere in the middle slice, one for grey matter (GM) in the gyrus dentatus in the head of the right hippocampus where it was larger and one in the background without artefact (BG). Mean (<X>) and standard deviation ($\sigma_X$) were computed for each region. RC was defined as: 2*[<GM>-<WM>]/[<GM>+<WM>]. SNR was defined as: <GM>/$\sigma_{BG}$.

2.4.2 Prevalence of motion artifact

In order to evaluate the prevalence of motion artifacts, T2-weighted TSE HR interleaved acquired slabs were evaluated for the 37 healthy subjects acquired on NS_7T_32CH. Each acquired HR slab was rated for motion artifacts on the whole brain according to three levels: 1. no motion artefact; 2. small to medium motion artifact (motion-induced ghosting and/or blurring can be observed, but anatomical structures remains visible); large motion artifact (motion-induced ghosting and/or blurring alter the visibility of several anatomical structures). Each acquired HR slab was considered specifically in order to evaluate the evolution of motion artifact frequency along the acquisition session.

2.4.3 Performance of multi-slab registration and prevalence of between-slab shifts

The performance of multi-slab registration was evaluated on the final full HR volume built from T2-weighted TSE sequences, for both contiguous and interleaved slabs, for the three acquisition settings. The visibility of hippocampal inner structure was evaluated in the acquisition plane as before and the 3D-consistency was characterized in sagittal plane by assessing any visible inconsistency within the structure and comparing the overall shape with the LR acquisition.

In order to assess the prevalence of information loss due to between-slab shifts, each pair of complementary interleaved HR slabs was evaluated specifically for between-slab shifts along the main axis of the hippocampus, for the 37 subjects acquired on NS_7T_32CH and the nine subjects acquired on

CMRR_7T_16CH. To do so, each final full HR volume (one pair of slabs, no averaging) was visualized in the sagittal plane and compared with the LR acquisition. Antero-posterior "steps" were then searched in the final full HR volume for borders which should appear smooth as in the LR acquisition; in fact, these "steps" correspond to information redundancy on two consecutive slices, which is the dual consequence of information loss. In order to evaluate the consequences of these shifts for further analyses, both first and repeated pairs of slabs were analyzed, for the subset of subjects for whom repeated HR interleaved slabs were available.

The number of subjects for whom a HR full volume was finally available for reliable analysis was also evaluated, by combining checks on motion artifacts, between-slab shifts and registration performance.

3. Results

Acquisition pilot tests were compared in order to choose the best protocol for imaging the hippocampal inner structure in-vivo at 7T. First, the visibility of the inner structure of the hippocampus for T2-weighted TSE (both contiguous and interleaved) and T2*-weighted GRE contrasts was analyzed. Then between slab motion and registration performance were investigated.

3.1 Visibility of hippocampal inner structure

50 healthy subjects were used for this visual evaluation (37 acquired with NS_7T_32CH, nine acquired with CMRR_7T_16CH and four acquired with CMRR_7T_32CH). Figure 4 and Figure 5 illustrate the acquisition performed in, respectively, the head and the body of the hippocampus for T2-weighted TSE and T2*-weighted GRE acquisitions with NS_7T_32CH; the layers sparsely populated with neuronal bodies, such as SLRM, appear darker and provide a good insight of hippocampal inner structure. For T2-weighted TSE acquisitions, thickness ranged between one and four voxels for the SRLM, one and two voxels for the alveus and one and five voxels for the SP, thus validating the choice of ultra-high in-plane resolution. However, for T2*-weighted acquisitions, the appearance of SRLM varied with the echo time, and was less sharp for shorter echo times, such as TE1=16.4ms, than for longer echo times, such as

TE2=33.2ms (Figure 4). This variation also yielded variations in the apparent SLRM thickness (e.g. on the slice shown here, two voxels for TE1 and three voxels for TE2).

Figure 6 illustrates T2-weighted TSE acquisitions performed in the head and the body of the hippocampus with CMRR_7T_16CH (with averaging) and CMRR_7T_32CH. Hippocampal inner structure was also clearly visible in both cases thanks to the lower signal in the SLRM.

Figure 7 illustrates interleaved and consecutive T2-weighted TSE acquisitions with NS_7T_32CH for a test subject. Contiguous and interleaved acquisitions were independently compared by two raters for the 11 subjects without motion artefact among the 19 healthy subjects for whom both acquisitions were available. The two raters agreed that the interleaved acquisition provided a better visibility of the inner structure of the hippocampus for nine subjects. For the two remaining subjects, the raters rated either the two images are equivalent or one of them rated the two images as equivalent and the other the interleaved acquisition as providing better visibility. These results were further confirmed by computing RC and SNR for all the images acquired and combined for a representative subject with no motion artefact. RC (respectively SNR) ranged from 0.14 to 0.26 (resp from 25 to 28) for raw interleaved acquisitions, and was smaller for the contiguous acquisition with RC = 0.08 (resp SNR = 22). For recombined images without averaging (thus following registration), RC ranged from 0.18 to 0.23 (resp SNR from 40 to 41) for interleaved acquisitions while they remained lower for the contiguous acquisition, with RC = 0.07 (resp SNR = 35).

3.2 Prevalence of motion artefacts

Motion artefacts on T2-weighted TSE acquisitions are illustrated in Figure 8 for three levels of motion: no motion, medium motion and large motion. Motion was evaluated for 37 subjects acquired with the NS_7T_32CH configuration; for each subject, motion was evaluated independently for the four high resolution slabs. Results are detailed in Table 4.

No motion artefact was observed in 51% to 67% of the slabs and medium motion artefact in 19% to 38% of the slabs. Large motion artefacts thus only occurred in 11% to 19%, showing a large variability between the first and the repeated acquisitions. Interestingly, large motion artefacts occur more often for the first slabs.

3.3 Performance of multi-slab registration and prevalence of between-slab shifts

The registration of contiguous slabs was evaluated for 19 subjects on T2-weighted TSE acquisitions acquired with NS_7T_32CH. It proved to allow accurate registration of the two contiguous slabs in all 19 subjects, even with motion artefact (Figure 9). Subject's movements between two slabs generated a shift between the two parts of the SRLM (Figure 9.a); this shift was corrected after registration (Figure 9.b) and the intensity was properly corrected by the masks in order to provide a 3D-consistent high resolution volume of the hippocampus (Figure 9.c). There was no signal loss due to antero-posterior shift.

The registration of interleaved slabs was evaluated on 37 subjects with T2-weighted TSE slabs acquired with NS_7T_32CH and nine subjects with T2-weighted TSE slabs acquired with CMRR_7T_16CH. When no large antero-posterior shift could be observed, the initial shift (Figure 10.a and Figure 11.a) was corrected after registration (Figure 10.b and Figure 11.b) and the intensities were homogenized using the corresponding phantoms (Figure 10.c and Figure 11.c).

When a large antero-posterior between-slab shift was observed (Figure 12), information is intrinsically lost, as shown when comparing the same subject acquired with an antero-posterior between-slab shift (first repetition) and with no antero-posterior between-slab shift (second repetition). In fact, for the first repetition, there was little difference between odd and even slices (redundancy), and it appears as obvious on the final result (Figure 12.b) as compared to the thorough consistent signal retrieved with the second repetition (Figure 12.c). Note that the low resolution volume appears very useful to detect such occurrences (Figure 12.a).

For interleaved T2-weighted TSE acquisitions with NS_7T_32CH, antero-posterior between-slab shift was detected in nine subjects out of 37 (24%) in the first repetition and for eight subjects out of 37 (22%) in the second repetition. Antero-posterior between-slab shift was observed in both repetitions only in two subjects among the 37 (5%).

For interleaved T2-weighted TSE acquisitions with CMRR_7T_16CH, antero-posterior between-slab shift was detected in one subject out of nine (11%) in the first repetition and for one subject out of nine (11%) in the second repetition. Antero-posterior between-slab shift was observed in no subject for both repetitions. Note that the between-slab shift can also be easily detected during the acquisition session.

Overall, only two subjects among the 46 acquired with NS_7T_32CH and CMRR_7T_16CH showed an antero-posterior between-slab shift in both repetitions and had to be excluded from further analyses. The remaining 44 subjects did not contain large motion artifacts on both repetitions and could thus be kept for further analyses. Thus, 44 subjects among the 46 acquired were of sufficient quality for further analyses of hippocampal inner structure (96% of the acquired datasets).

## 4. Discussion

This study aimed at building a thorough procedure for robust in-vivo imaging of hippocampal inner structure at 7T. The final procedure is based on multi-slab acquisitions which make it possible to reduce acquisition time for each slab, together with a robust registration framework. In order to better address the specific constraints related to the inner structure of the hippocampus, a T2-weighted TSE acquisition was chosen, acquired perpendicular to the main axis of the hippocampus with ultra-high in-plane resolution and larger slice thickness. Extensive visual evaluation on 46 subjects acquired with two 7T systems proved that the proposed acquisition setting makes it possible to obtain precise visualization of details of the inner structure on the whole hippocampus. It was thus efficient and robust, 96% of the subjects being of sufficient quality to be further analyzed.

Comparing T2 and T2* weightings was motivated by their acknowledged performance in revealing the inner structure of the hippocampus [18–24]. Even though T1 contrast was investigated for the same purpose at 7T [18, 22] and at lower field strengths [11, 29], the visibility of the SLRM for this contrast was not sufficient for further quantitative analyzes, and was thus not considered here. Although the visibility of the SRLM was better for T2*-weighted GRE acquisitions with large TE than for T2-weighted TSE acquisitions, its apparent thickness highly depended on the TE value for T2*-weighted GRE acquisitions; this was an issue for segmentation as delineation was made uncertain. Moreover, T2*-weighted GRE acquisitions were altered by distortion and susceptibility artifacts that made the registration of multi-slab acquisitions undependable. Finally, T2*-weighting is a mixture of T2 and susceptibility-based weightings, making it more sensitive to blood vessels. Therefore, T2*-weighted images of the hippocampal area are more difficult to analyze, due to the dense vessel concentration. T2-weighting was thus considered as more robust for further quantitative analyses in patients, and the high sensitivity of the 32 channel head coils makes it possible to minimize the SAR-related issues usually related to T2-weighted acquisition.

Hippocampal geometry in the body makes it possible to use high in-plane resolution with larger slice thickness, although this configuration may not be ideal for defining hippocampal inner structure in the head and the tail. Nevertheless, reducing slice thickness while keeping sufficient in-plane resolution would result in less signal and thus lower visibility of hippocampal inner structure. Furthermore, data presented in this paper showed good visibility in the head and lower visibility only in the three or four most posterior slices.

Data shown here were obtained with either contiguous or interleaved 2D acquisitions. A comparison performed by two raters showed that interleaved 2D acquisitions provided a better visibility of the inner structure of the hippocampus than contiguous 2D acquisitions. This finding was confirmed by contrast and SNR measurements. 3D acquisitions might be considered as a mean to obtain more signal and thus move to a more isotropic resolution. Nevertheless, acquisition tests with 3D-T2-SPACE showed that it was highly sensitive to subject's movement and B1 field inhomogeneities, resulting in a large signal loss in the temporal lobe.

2D T2-weighted or T2*-weighted acquisition procedures at 7T were previously proposed for imaging hippocampal inner structure perpendicular to the main axis of the hippocampus [19, 22, 24]. Even though resolutions of about 300 μm in-plane were proposed [19, 22, 24], the antero-posterior coverage was limited, either through larger slice thickness [22] or a gap between slices [24], or with 1mm slice thickness but partial coverage of the hippocampus with a 14 minutes scan [19]. Our team already proposed a multi slab acquisition procedure to reduce motion [23] but no methodological details on the registration method and no assessment regarding motion sensitivity of contiguous or interleaved acquisitions were given.

The registration method proved robust on both interleaved and contiguous TSE T2-weighted acquisitions for data acquired on three MRI systems. However, possible signal loss due to antero-posterior between-slab shift could remain an issue for interleaved acquisition. The evaluation provided here on 46 subjects showed that such motion is unlikely to occur for both repetitions, if two repetitions are acquired. Furthermore, interleaved acquisitions were shown to allow for a better visibility than contiguous ones, which was expected due to the cross-talk phenomenon on T2-weighted images. The final procedure thus embedded repeated interleaved T2-weighted slabs together with a full slab at lower resolution for precise registration.

Overall, motion artefacts on high resolution slabs were shown to be less frequent in slabs acquired in the middle of the acquisition slot (second slab of the first repetition and first slab of the second repetition) compared to first and last slabs. This could be explained by two phenomena: first, subjects do not feel their own motion, and need a "training" before being able to grasp the very small movement that can weaken image quality (large motion artifacts are more frequent for the first slab); second, keeping motionless in the scanner engenders strains in the neck and overall body of the subjects who are not able to control motion after some time (medium motion artifact were more frequent in the last slab). This emphasizes the need of a precise procedure to optimize acquisition quality while ensuring feasibility in patients. We did not directly address the issue of image quality with respect to sequence duration. Nevertheless, this issue is not easy to address, as the overall time spent in the scanner before each sequence has to be taken into account, and the relationship between motion and time in the scanner is not straightforward and rather subject dependent.

The time reduction provided for each acquisition will provide a considerable reduction of motion artifacts especially for patients. Therefore, the proposed procedure is of a great interest for many neurological disorders like Alzeihmer's disease [27] or temporal lobe epilepsy [23] for which hippocampal subfields are known to be affected differently[23, 27]. In fact, such procedure would allow less rejection of subjects (due to motion artifacts) in a given database and would provide a more precise delineation of hippocampal subfields. This would allow having stronger statistics and more specific characterization of the disease. Moreover with such a procedure, the inner structure of the whole hippocampus could be delineated which would provide a more complete characterization of the disease than previous studies which focused only on parts of the hippocampus [27].

## 5. Conclusion

We have proposed an efficient procedure to address the specific constraints of imaging hippocampal inner structure in-vivo at 7T relying only on manufacturer standard sequences. Once adequate contrast and resolution had been chosen to ensure sufficient visibility of hippocampal inner structure, a procedure was proposed to reduce the time per acquisition and thus motion artefacts. An evaluation of the feasibility and robustness of the overall protocol was also proposed. The final procedure embedded multi-slab acquisition combined with a robust registration framework to yield a final 3D-consistent volume. The registration framework is available for download and could be applied to any other acquisition setting based on interleaved slabs acquisitions. Furthermore, the evaluation on three acquisition settings gives an insight on the robustness of the protocol with respect to the technical configuration used for data acquisition.

**Funding:**

This work was supported by ANR (project HM-TC, grant number ANR-09-EMER-006), France Alzheimer Association (project IRMA7), by the program "Investissements d'avenir" (grant number ANR-10-IAIHU-06) and by the CATI project (Fondation Plan Alzheimer)

**Conflict of interest:**

The authors declare that they have no conflict of interest

**Ethical approval:**

All procedures performed in studies involving human participants were in accordance with the ethical standards of the institutional and/or national research committee and with the 1964 Helsinki declaration and its later amendments or comparable ethical standards.

**Informed consent:** Informed consent was obtained from all individual participants included in the study.

**Author's contribution**

Protocol/project development MC, OC, AV, PFVM, CP, LHP, SL, LM

Data Collection or management MC, LM, JG, AV, PFVM, JS, TH

Data analysis LM, MC, AV

|  | NS_7T_1TX_32RX | CMRR_7T_16CH | CMRR_7T_32CH |
| --- | --- | --- | --- |
| machine | 7T Siemens Healthcare Magnetom (Erlangen, Germany) | 7T Siemens Healthcare Magnetom (Erlangen, Germany) | 7T Siemens Healthcare Magnetom (Erlangen, Germany) |
| gradient | head gradient AC84 with max slew rate 333 $(T/m).s^{-1}$ and max gradient amplitude 80mT/m | Body gradients AS095DS with max slew rate 200 $(T/m)s^{-1}$ and max gradient amplitude 38mT/m | body gradients AC72 with max slew rate 200 $(T/m).s^{-1}$ and max gradient amplitude 70mT/m |
| coil | a Nova Medical (Wilmington, Massassuchet, USA) single transmission RF coil and 32 receiving channel coil arrays | a home-made 16 channel transceive coil array | a Nova Medical (Wilmington, Massassuchet, USA) single transmission RF coil and 32 receiving channel coil arrays |

**Table 1 Acquisition settings used in the two research centers: (NS_7T_1TX_32RX) used in NeuroSpin and (CMRR_7T_16CH, CMRR_7T_32CH) used in CMRR.**

|  | NS_7T_32CH | | | | | | | | | | |
| --- | --- | --- | --- | --- | --- | --- | --- | --- | --- | --- | --- |
|  | nb | Resolution (mm x mm x mm) | Number of slices/slab | Acquisition time min/slab | TR (ms) | TE (ms) | A ° | FoV (mm) | Acquisition matrix | Bandwidth Hz/pixel | Turbo factor |
| T2w TSE contiguous | 19 | 0.3x1.2x0.3 | 23 | 5 | 5000 | 82 | 60 | 173x173 | 576x576 | 121 | 9 |
| T2w TSE interleaved | 37 | 0.3x1.2x0.3 1.2mm gap | 23 | 5 | 5000 | 82 | 60 | 173x173 | 576x576 | 121 | 9 |
| T2w TSE low resolution | 37 | 0.3x1.2x0.6 | 46 | 4:50 | 8000 | 80 | 60 | 173x173 | 311x576 | 121 | 9 |
| T2*w GRE | 37 | 0.3x1.2x0.3 | 45 | 12 | 791 | 16/33 | 65 | 173x173 | 576x576 | 70/70 |  |
|  | CMRR_7T_16CH | | | | | | | | | | |
| T2w TSE interleaved | 9 | 0.25x1.2x0.25 1.2mm gap | 30 | 5:04 | 5830 | 64 | 60 | 119x130 | 472x512 | 175 | 9 |
| T2w TSE low resolution | 9 | 0.25x1.2x0.5 | 60 | 5:08 | 11800 | 64 | 60 | 119x130 | 236x512 | 175 | 9 |
|  | CMRR_7T_32CH | | | | | | | | | | |
| T2w TSE interleaved | 4 | 0.25x1.2x0.25 1.2mm gap | 16 | 5:37 | 6000 | 55 | 120 | 130x130 | 512x512 | 174 | 9 |
| T2w TSE low resolution | 4 | 0.25x1.2x0.5 | 62 | 5:37 | 12000 | 54 | 120 | 130x130 | 256x512 | 174 | 9 |

**Table 2: acquisition parameters for NS_7T_32CH, CMRR_7T_16CH and CMRR_7T_32CH. Nb: number of subjects,TR: Repetition Time, TE: Echo Time, A : refocusing angle, FoV: Field of View.**

| |
|---|
| Localizer |
| Specific localizer for positioning hippocampal slabs |
| Field map |
| T2w TSE interleaved, slab 1 |
| T2w TSE interleaved, slab 2 |
| T2*w GRE |
| T2w TSElow resolution |
| T2w TSE interleaved, slab 1 |
| T2w TSE interleaved, slab 2 |
| T2w TSE contiguous, slab 1 |
| T2w TSE contiguous, slab 2 |
| MP RAGE |

**Table 3: standard acquisition protocol. The gray cells correspond to optional sequences**

|  | First repetition | | Second repetition | |
| --- | --- | --- | --- | --- |
|  | Slab1 | Slab2 | Slab1 | Slab2 |
| No motion artefact | 59% | 54% | 67% | 51% |
| medium motion artefact | 22% | 30% | 19% | 38% |
| Large motion artefact | 19% | 16% | 14% | 11% |

**Table 4. Percentage of the occurrence of the three levels of motion artefacts for the different slabs in the first and second repetitions of 2D TSE interleaved acquisitions.**

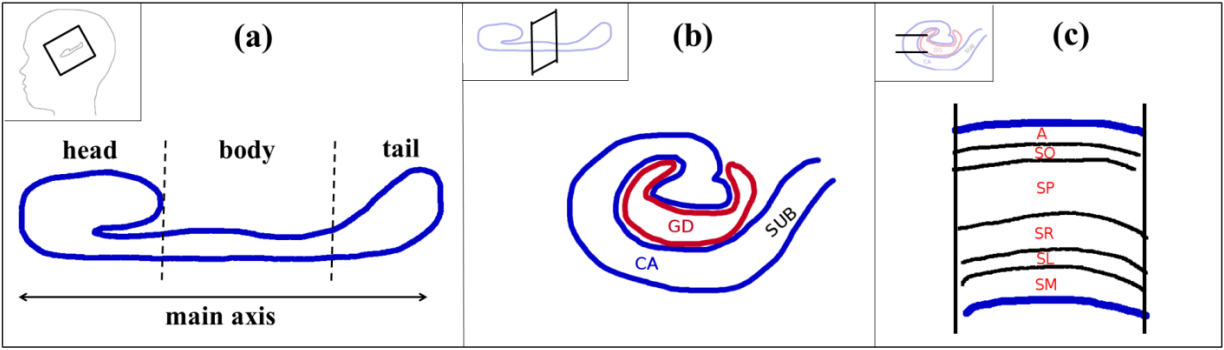

**Figure1: Anatomy of the hippocampus. (a) Sagittal view of the hippocampus. (b) CA and GD enrollment. (c) CA composition: A: alveus, SO: stratum Oriens, SP: stratum pyramidale, SR: stratum radiatum, SL: stratum lacunosum, SM: stratum moleculare.**

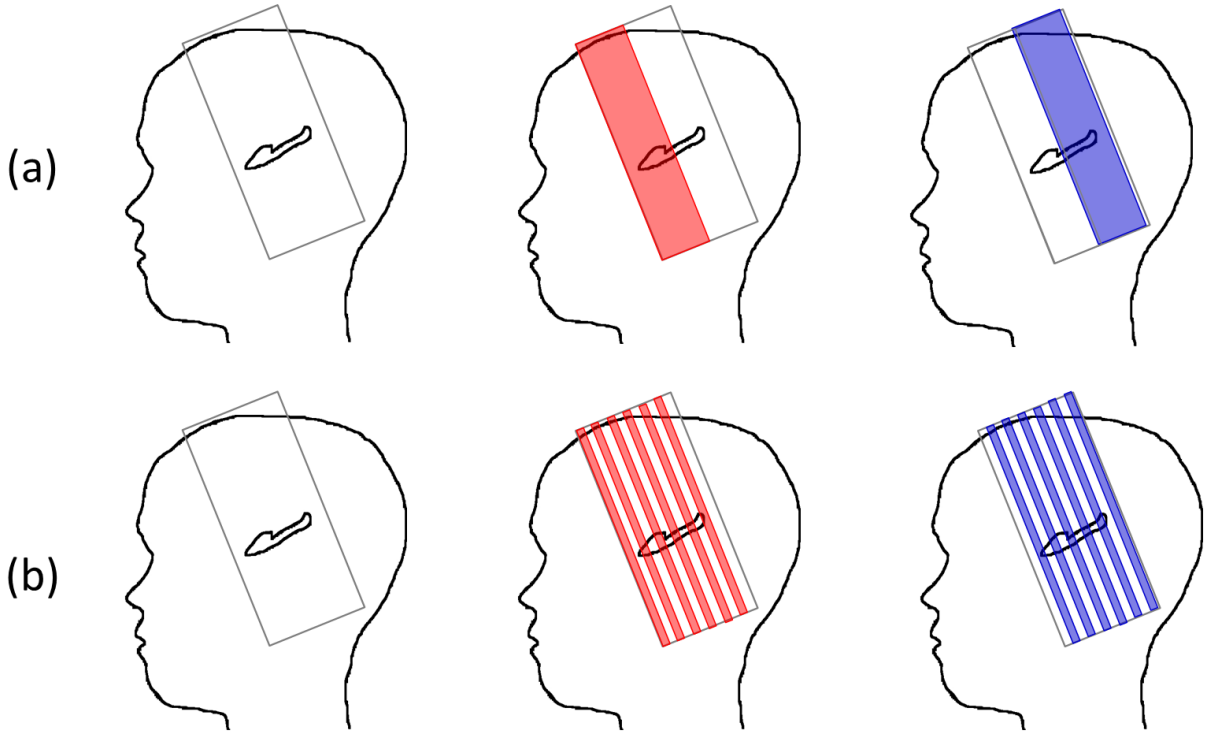

Figure 2: Multi-slab acquisitions for the hippocampus inner structure imaging: (a) contiguous slabs acquisition, (b) interleaved slabs acquisition. The first slab is represented in red and the second slab in blue.

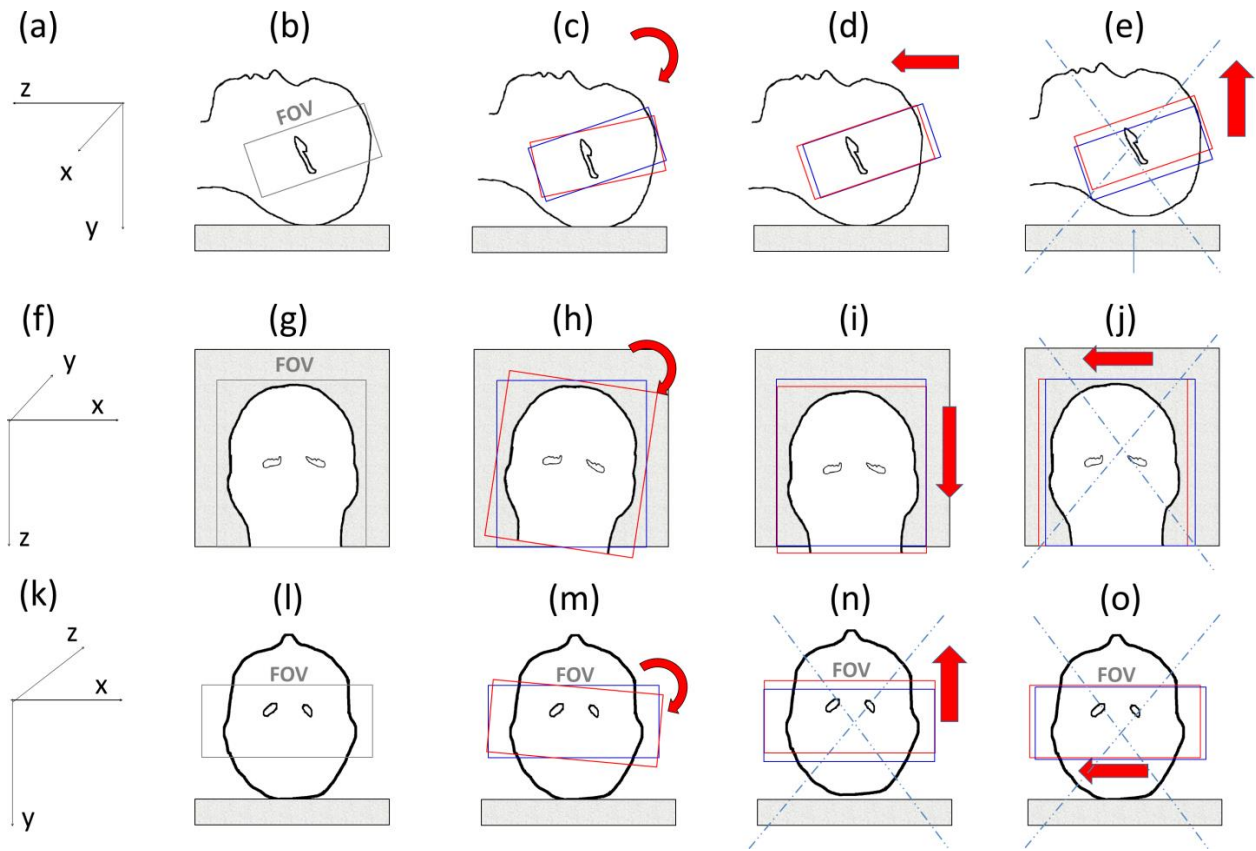

**Figure 3: Possible motion inside the MRI system (top row: sagittal view, bottom row: coronal view) (a) xyz frame; (b) position of the first slab perpendicular to the main axis of the hippocampus; (c) possible rotation along the x axis; (d) possible translation along the z axis; (e) impossible translation along the y axis; (f) xyz frame; (g) position of the first slab; (h) possible rotation along y axis; (i) possible translation along z axis; (j) impossible translation along x axis (first slab in red, second slab in blue); (k) xyz frame; (l) position of the first slab; (m) possible rotation along z axis; (n) impossible translation along y axis; (o) impossible translation along x axis**

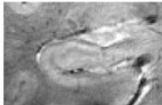

**Figure 4: Visibility of the whole head of the hippocampi for four subjects and for T2\*-weighting TE1= 16.41ms; T2\*-weighting TE2=33.22ms; T2-weighting high-resolution slab.**

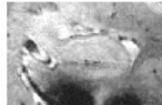

**Figure 5: Visibility of the whole body of the hippocampi for four subjects and for T2\*-weighting TE1= 16.41ms; T2\*-weighting TE2=33.22ms; T2-weighting high-resolution slab.**

| | Subject1_16CH | Subject2_16CH | Subject3_32CH |
|---|---|---|---|
| head | 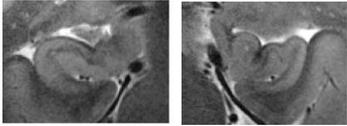 | 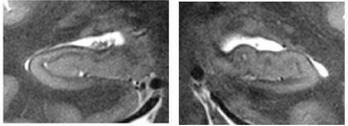 | 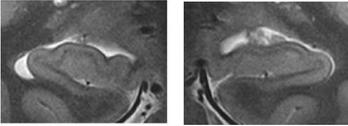 |
| body | 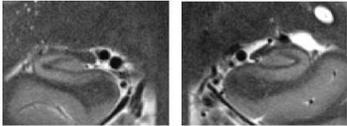 | 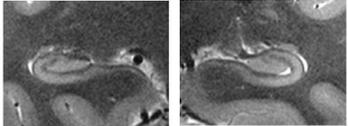 | 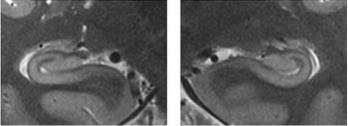 |

**Figure 6**: T2w TSE acquisition performed on the head and body of the hippocampus for CMRR_7T_16CH on two test subjects (with averaging) and for CMRR_7T_32CH on one test subject (without averaging).

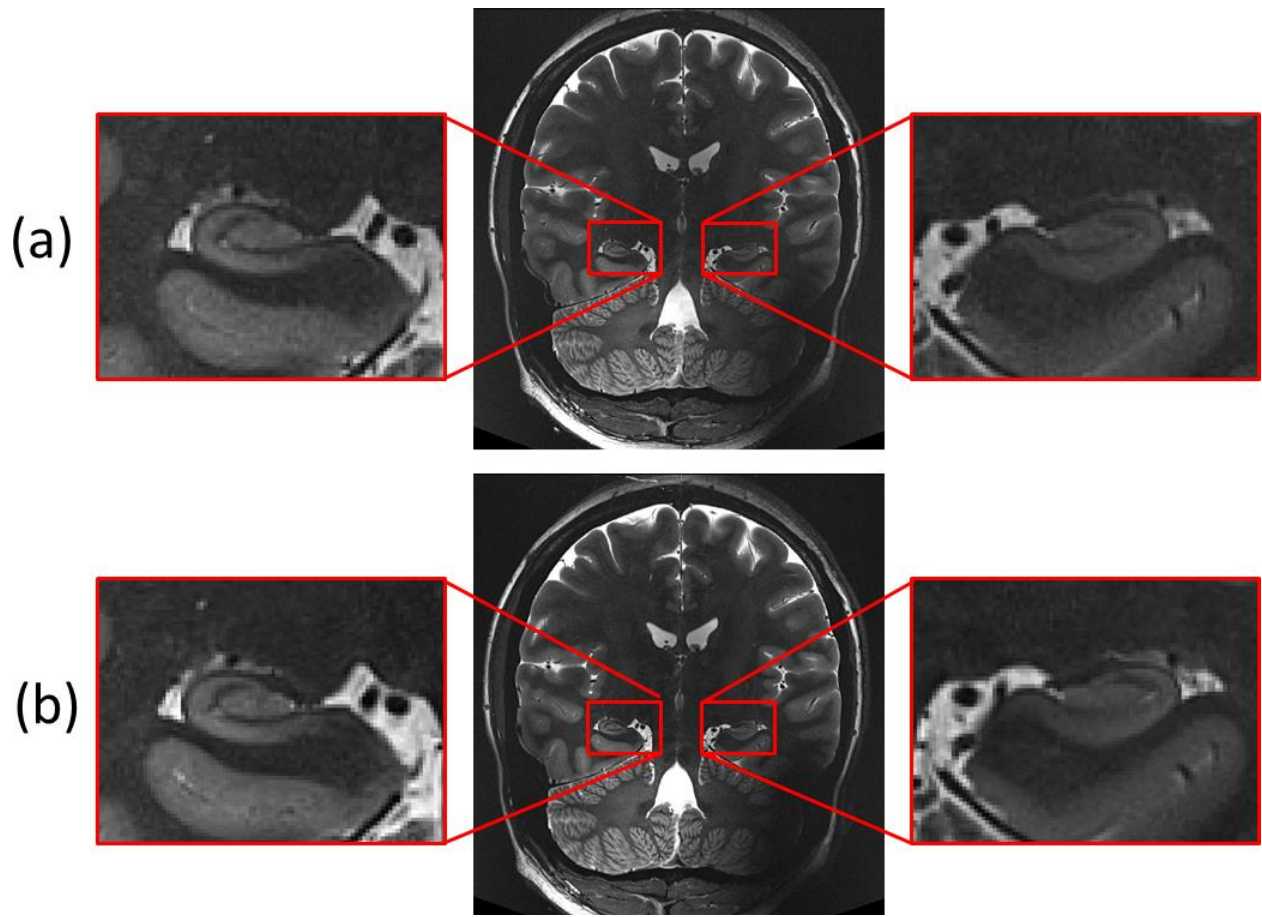

**Figure 7: Comparison of (a) contiguous acquisition, (b) interleaved acquisition for the same subject**

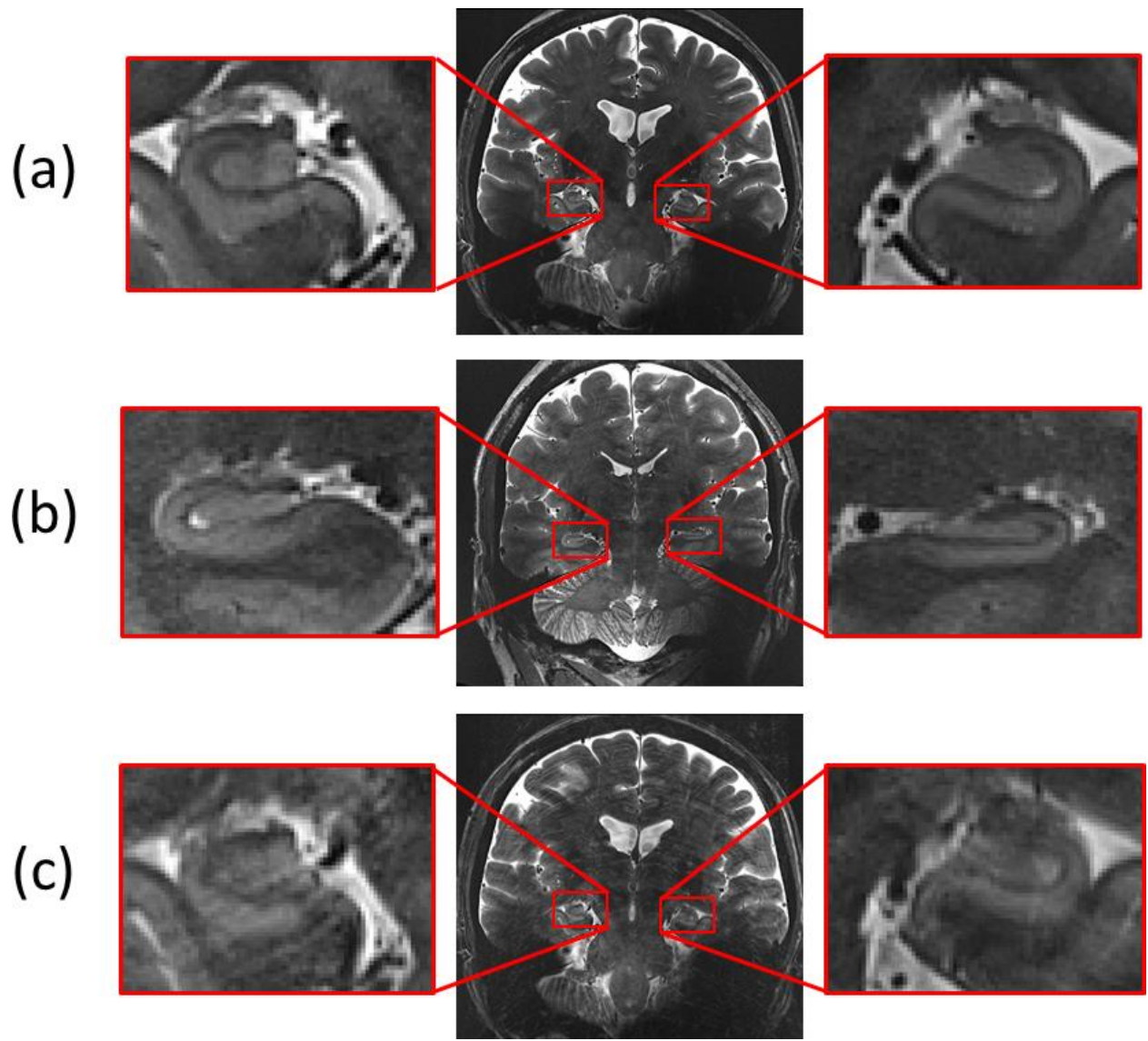

**Figure 8: motion artefact: 3 levels of motion artefact: (a) no motion artefact, (b) medium motion artefact, (c) large motion artefact.**

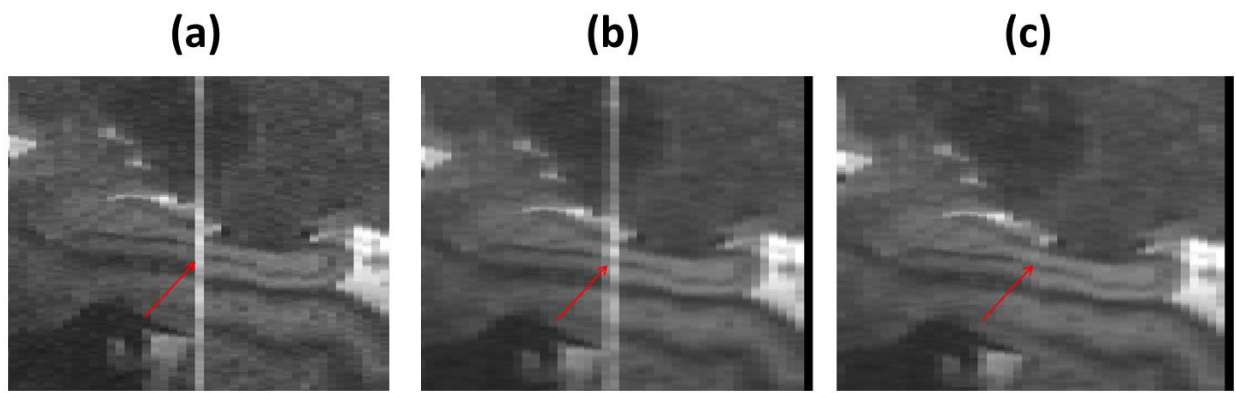

**Figure 9: Contiguous slabs registration for a subject acquired on NS_7T_32CH. (a) Sum of unregistered slabs. (b) Sum of registered slabs. (c) After intensity homogenization.**

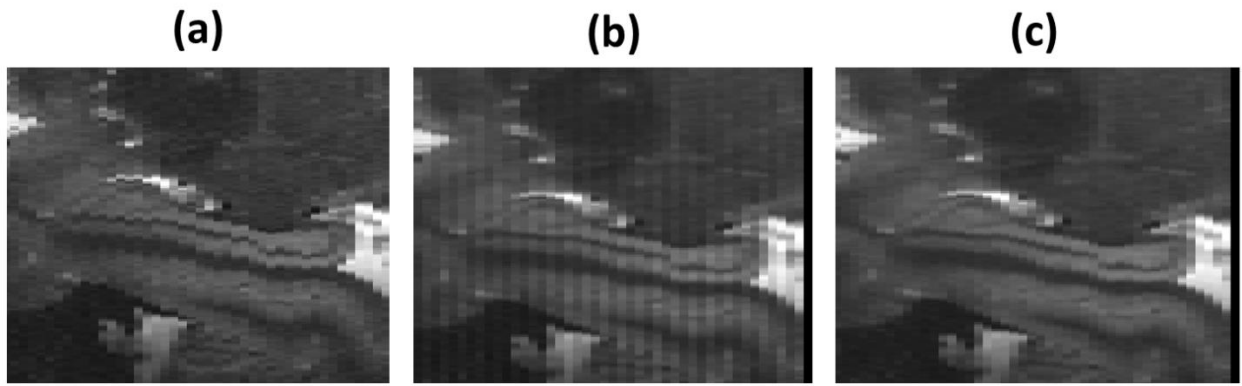

**Figure 10: Interleaved slabs registration for a test subject acquired on NS_7T_32CH. (a) Sum of unregistered slabs. (b) Sum of registered slabs. (c) After intensity homogenization.**

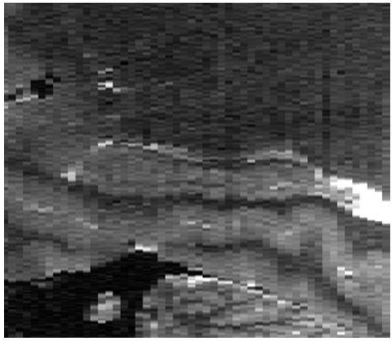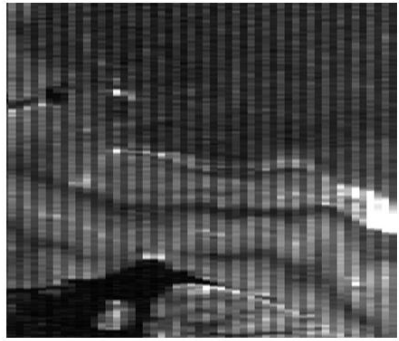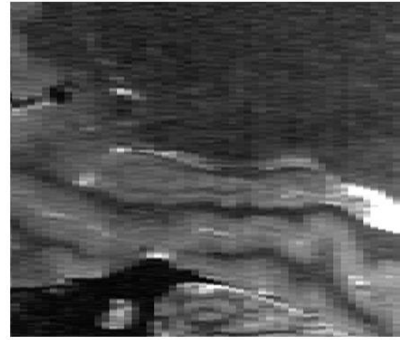

Figure 11: Interleaved slabs registration for a test subject acquired on CMRR_7T_16CH. (a) Sum of unregistered slabs. (b) Sum of registered slabs. (c) After intensity homogenization.

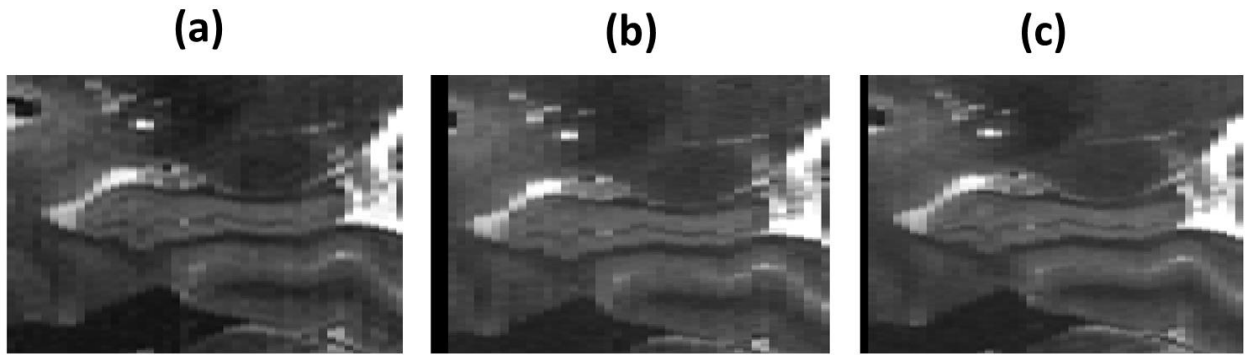

**Figure 12. Between-slab antero posterior shift for interleaved acquisition. (a) low-resolution reference scan (b) High resolution final 3D-consistent slab for the first repetition with antero-posterior between slab shift(c) High resolution final 3D-consistent slab for the first repetition with no antero-posterior between slab shift.**